\documentclass{article}

\PassOptionsToPackage{inline}{enumitem}

\def\pkneuripstrack{position}
\usepackage[arxiv,nonatbib]{styles/paperkit}

\usepackage{common}

\usepackage[style=numeric-comp,backend=biber,natbib=true,giveninits=true,%
            maxbibnames=8,sorting=nyt]{biblatex}
\papertitle[The Greatness of Science Cannot Be Planned: Agentic Auto-Research is Fuzz Testing]{The Greatness of Science Cannot Be Planned:\\Agentic Auto-Research is Fuzz Testing}
\paperrunningtitle{Agentic Auto-Research is Fuzz Testing}
\paperauthors{%
  Yifeng~He\textsuperscript{1,2}\quad
  Jicheng~Wang\textsuperscript{1}\quad
  Yinzhe~Zhao\textsuperscript{3}\textsuperscript{\ensuremath{\dagger}}\quad
  Chengyang~Shi\textsuperscript{2,4}\quad
  Jiachen~Liu\textsuperscript{2}\quad
  Hao~Chen\textsuperscript{5}
}
\paperaffiliations{%
  \textsuperscript{1}University of California, Davis \quad
  \textsuperscript{2}ARA Lab \quad
  \textsuperscript{3}Zhejiang University\\
  \textsuperscript{4}University of Michigan \quad
  \textsuperscript{5}The University of Hong Kong}
\papercorrespondence{%
  \href{mailto:yfhe.cs@gmail.com}{yfhe.cs@gmail.com};
  \href{mailto:amber@ara-commons.com}{amber@ara-commons.com};
  \href{mailto:chenho@hku.hk}{chenho@hku.hk}
  }
\paperlogo{}{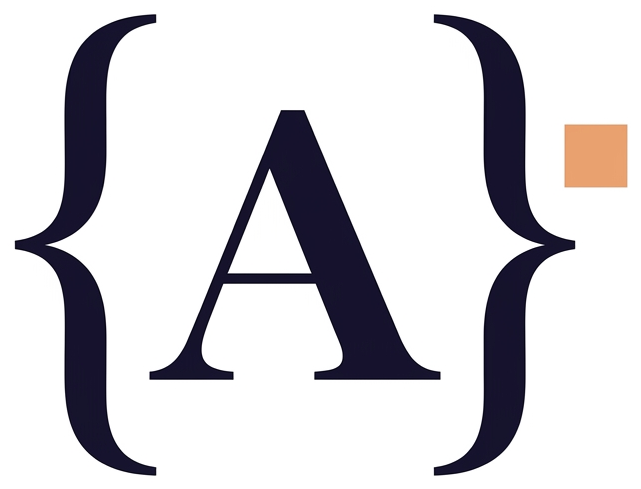}
\paperlogoheight{20pt}
\paperabstract{%
Agentic auto-research is emerging, but most systems treat scientific discovery as goal-oriented optimization against a final benchmark.
This paradigm rewards a sparse final verdict and ignores the exploration that precedes it.
When agents optimize only the final score, they overfit to the test conditions and sample blindly rather than search.
Within a declared research problem, a research agent and a greybox fuzzer for software analysis face the same sparse feedback.
A fuzzer rarely finds a bug directly, but coverage makes partial progress observable on every execution.
Fuzzers use that dense signal to mutate inputs and allocate effort, rather than merely rank completed runs.
Auto-research needs the same two capabilities.
First, each experiment must expose a cheap, dense signal of epistemic progress before final scientific validation is available.
Second, that signal must determine the next intervention so the agent searches rather than repeatedly samples.
Because the progress signal provides guidance rather than a final verdict, final validation must still evaluate claims using evidence protected from adaptive reuse.
We propose controlled tests to determine whether candidate signals predict validated progress,
whether feedback-directed search yields more validated discoveries per unit cost than repeated sampling,
and whether protected validation reduces false discoveries. 
In a simulated physics environment, an AI research agent that tracks its intermediate epistemic progress discovers a hidden physical law.
Optimization-driven baselines fail because they repeatedly sample and overfit to their existing data instead of probing unfamiliar regimes.
Feedback architecture, not generation capacity, is the central bottleneck in auto-research.
}

\makeatletter
\newcommand{\pkblankfootnote}[1]{\begingroup\xdef\@thefnmark{}\@footnotetext{#1}\endgroup}
\makeatother

\begin{document}

\makepaperheader
\pkblankfootnote{\textsuperscript{\ensuremath{\dagger}}Work completed while
  interning at UC Davis.}

\section{Introduction}

\begin{figure*}[!t]
  \centering
  \includegraphics[width=.8\linewidth]{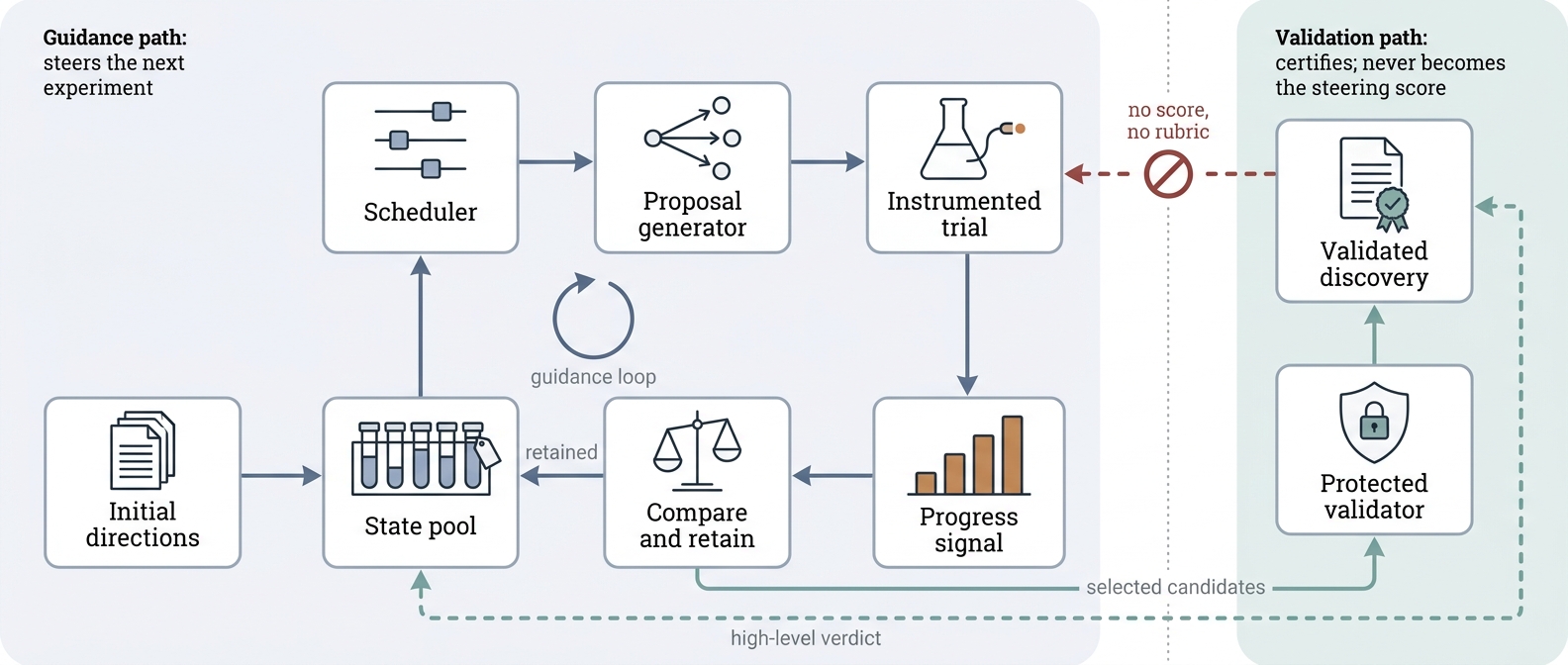}
  \caption{Auto-research as feedback-directed search. Initial directions enter a
  state pool. A scheduler selects a state, the proposer generates candidates, and
  experiments expose intermediate signals. Comparison retains promising results
  as new states and closes the loop. A protected validator certifies discoveries
  and returns a high-level verdict to the state pool, never a score or a rubric
  the search could optimize.}
  \label{fig:loop}
\end{figure*}

Agentic auto-research has emerged as an approach to conduct science day and night, yet current system designs predominantly treat scientific discovery as a goal-oriented optimization problem against a final benchmark. Researchers build systems that tirelessly propose hypotheses, write code, and evaluate results simply to maximize a target metric \citep{lu2024aiscientist,yamada2025aiscientistv2,gottweis2025coscientist}.
Optimizing for a sparse final score ignores the exploration in between: agents overfit to the test conditions and sample blindly rather than search. In \autoref{fig:loop}, we preview our position: auto-research should use intermediate signals to retain promising states and schedule new experiments, while reserving scientific claims for protected validation.

The fundamental nature of science, however, is exploring the unknown, not simply winning a benchmark. Science maps the boundaries of ignorance and continuously updates our cognitive model of reality. Every dead end and failed experiment provides data to sharpen this understanding. Treating science purely as an optimization problem falls short because discovery cannot be achieved by blindly generating proposals to push a number up. Without intermediate epistemic progress, agents are merely guessing. Empirical results show that repeated sampling yields diminishing returns, as the fraction of problems solved grows only log-linearly with sample count \citep{brown2024monkeys}. Scientific evaluations often require long computations, expert labor, replication, or physical experiments. Instead of merely sampling for a final success, agents must extract dense epistemic progress from the environment to guide their next steps.

Fuzz testing offers a compelling solution for agentic auto-research. An autonomous research agent generates a candidate experiment, executes it, observes feedback, and chooses what to try next. Coverage-guided greybox fuzzing runs the same control loop \citep{afl,manes2021fuzzing}. Both domains receive sparse feedback on their ultimate goals. A fuzzer rarely finds a bug on any given run, just as a researcher rarely makes a breakthrough in a single experiment. The design principle of fuzzing is to mutate a seed input, execute the target program under lightweight instrumentation, and observe the intermediate code coverage. This coverage makes partial progress observable on every execution. Modern fuzzers use this dense, intermediate signal to adaptively schedule mutations, target unsolved constraints, and allocate effort \citep{fioraldi2020aflpp}. Research does not literally crash like software, but the same principle applies: \emph{an expensive search needs observable progress and a policy that acts on it}. By \emph{instrumented}, we mean equipped to expose intermediate feedback during execution. A fuzzer inserts probes at program branches; a scientist equips an experiment to measure intermediate variables. Both define these observations before execution so that the search can react to intermediate behavior, not only the final outcome.

\paragraph{Lesson One: Make Research Progress Observable.}
Final validation usually gives sparse feedback: most experiments do not establish a finding. Auto-research therefore needs an intermediate signal that reports partial progress on each candidate before final validation is available. The research analogue of coverage should track epistemic progress, such as whether an experiment rules out an explanation, locates a boundary, sharpens a prediction, or changes which experiment should run next. This signal is not an early verdict that the current candidate is correct. It exposes how the campaign's state changed and whether that change may help future search.

\paragraph{Lesson Two: Use Feedback to Search.}
Once useful guidance exists, the next experiment should depend on what previous experiments revealed. Fuzzers use execution feedback to mutate promising seeds, target unsolved constraints, and allocate energy across regions. A research agent should likewise use feedback to choose its next intervention, rather than only rank completed samples. This distinction becomes more important when each evaluation is expensive, noisy, or slow.

\paragraph{Lesson Three: Separate Exploration Signals from Final Validation.}
Auto-research requires a strict separation between exploring hypotheses and validating discoveries. While an intermediate progress signal helps an agent decide where to search next, it cannot serve as the final proof of success. Because the agent adaptively optimizes this signal during search, a separate validator must evaluate the final claims against independent, protected evidence to prevent overfitting. This mirrors the division of roles in fuzzing: coverage guides the fuzzer toward new program behavior, but a separate crash or sanitizer oracle determines whether a vulnerability actually exists \citep{manes2021fuzzing,fioraldi2020aflpp,barr2015oracle}.

\paragraph{Our position.}
\textbf{Auto-research should operate as instrumented, feedback-directed search.} Within a declared research problem, each experiment should expose a cheap, dense measure of epistemic progress; the search policy should use that measure to choose its next intervention; and final validation, protected from adaptive reuse of the progress signal, should determine what counts as a discovery. This position makes three separable predictions. First, candidate progress signals should predict protected scientific outcomes and improve how a fixed validation budget is allocated. Second, with the proposer, task, validator, and budget fixed, feedback-directed search should yield more validated discoveries per unit cost than repeated sampling. Third, with the search trace fixed, protected validation should reduce false discoveries relative to reporting the optimized proxy. These predictions specify how feedback should guide search; they do more than restate that discovery involves search.

\section{The Research Loop Is the Fuzzing Loop}

\begin{figure}[t]
  \centering
  \resizebox{\columnwidth}{!}{%
  \begin{tikzpicture}[x=1cm,y=1cm]
    \tikzset{cycbox/.style={figbox,minimum width=15.5mm,minimum height=8.5mm,
      inner sep=1pt,font=\scriptsize,text=figink,align=center}}

    \node[figlabel,anchor=west,font=\scriptsize\bfseries] at (0.05,4.30) {(a)};
    \node[cycbox] (ap) at (1.35,4.30) {Propose};
    \node[cycbox] (ai) at (3.35,4.30) {Implement\\and run};
    \node[cycbox,fill=figblood!8] (ae) at (5.35,4.30) {Evaluate};
    \draw[figflow] (ap) -- (ai);
    \draw[figflow] (ai) -- (ae);
    \draw[figflow] (ae.north) .. controls +(0,0.55) and +(0,0.55) .. (ap.north);
    \node[figannot,anchor=south,fill=white,inner sep=1pt] at (3.35,5.22)
      {one score steers and certifies};

    \node[figlabel,anchor=west,font=\scriptsize\bfseries] at (0.05,2.45) {(b)};
    \node[cycbox] (bp) at (1.35,2.45) {Propose};
    \node[cycbox] (bi) at (3.35,2.45) {Implement\\and run};
    \node[cycbox,fill=figclay!12] (bs) at (5.35,2.45) {Progress\\signal};
    \node[cycbox,fill=figteal!12] (bv) at (5.35,1.10) {Protected\\validation};
    \draw[figflow] (bp) -- (bi);
    \draw[figflow] (bi) -- (bs);
    \draw[figflow] (bs.north) .. controls +(0,0.55) and +(0,0.55) .. (bp.north);
    \draw[figcert] (bs) -- (bv);
    \node[figannot,anchor=south,fill=white,inner sep=1pt] at (3.35,3.37)
      {steers the next proposal};
    \node[figannot,anchor=east,align=right] at (4.40,1.10)
      {certifies;\\never steers};
  \end{tikzpicture}}
  \caption{Splitting the evaluate box. (a) The common research cycle
  scores each experiment with a single evaluation that both steers the search
  and certifies the result. (b) Fuzzing separates the two roles: a cheap
  progress signal steers the next proposal, while protected validation
  certifies discoveries and never feeds the search.}
  \label{fig:cycle}
\end{figure}
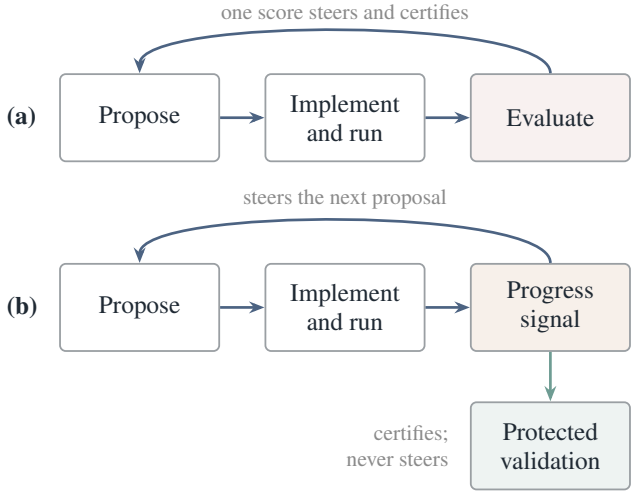

Science and engineering share one iterative cycle: propose an
experiment, implement and run it, evaluate the outcome, and let that outcome
shape the next proposal. Auto-research systems race to automate this cycle end
to end, and automation does drive down iteration time. Most systems,
however, draw evaluation as a single box whose one score both steers the search
and certifies the result. In \autoref{fig:cycle}, we contrast this design with
fuzzing's split: cheap coverage steers, and a protected oracle certifies. The
rest of this section maps the two loops component by component.

The correspondence concerns control structure, not literal equivalence. Each fuzzing component exposes a design decision that automated research systems also make. In \autoref{fig:loop}, we arrange these components as a loop; in \autoref{tab:analogy}, we compare them component by component. Prior work and data supply seeds; the LLM proposes mutations; experiments provide instrumented executions; and a scheduling policy allocates compute and validation budget across directions \citep{aflfast,aflgo}. The central mapping first asks what an execution can reveal before a rare success occurs, then asks how that observation changes the next candidate. A protected validator remains outside this guidance path and determines which candidates count as discoveries.

Research differs from a fixed program because it can revise its hypotheses and representations while it runs. Our analogy therefore covers search within a declared research problem, even when the search space is too large to enumerate. It does not cover the abductive act of changing what counts as the problem, a candidate, or a discovery. We return to this boundary in \autoref{sec:alternatives}.

\begin{table}[t]
  \caption{The analogy, component by component. The boldfaced rows identify the two central design choices: make intermediate progress observable and use it to direct search.}
  \label{tab:analogy}
  \centering
  \footnotesize
  \setlength{\tabcolsep}{4.5pt}
  \begin{tabular}{p{0.21\linewidth}p{0.25\linewidth}p{0.45\linewidth}}
    \toprule
    Fuzzing & Automated research & Transferable lesson \\
    \midrule
    Seed corpus & Prior results and literature & Seed quality bounds what search can reach \\
    Mutation operator & LLM proposer of experiments & Strict operators keep validity; many broaden reach \\
    \textbf{Coverage (guidance)} & \textbf{Intermediate progress signal (\emph{open})} & Expose useful progress on each experiment \\
    Crash / sanitizer (oracle) & A validated discovery & Check each relevant operation; positive verdicts are rare \\
    \textbf{Blind vs.\ targeted mutation} & \textbf{Repeated vs.\ directed search} & Let observations choose transformations and allocations \\
    Power schedule & Compute across directions & Prioritize under-explored, high-yield regions \\
    \bottomrule
  \end{tabular}
\end{table}

\section{Lesson One: Make Research Progress Observable}
\label{sec:signal}

\begin{figure}[t]
  \centering
  \includegraphics[width=\linewidth]{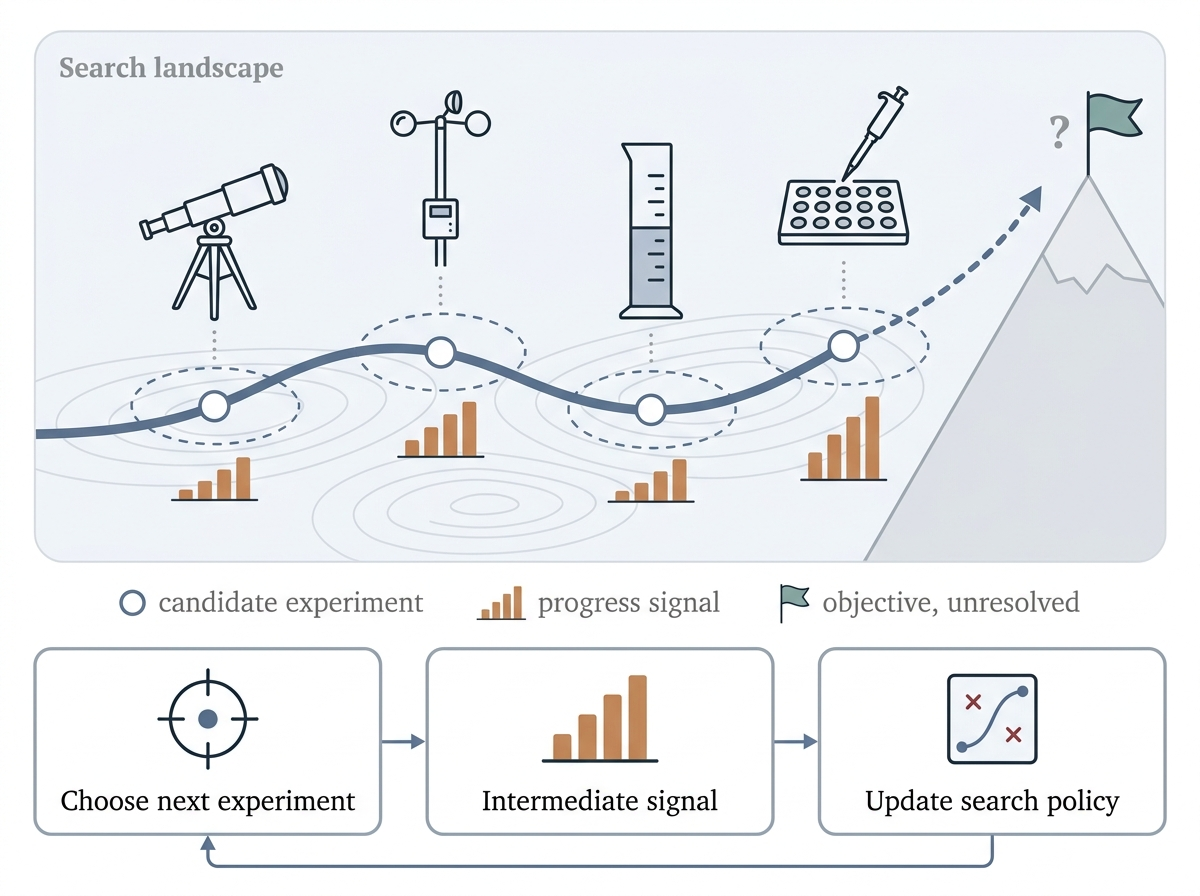}
  \caption{Intermediate signals make partial research progress observable before
  final validation. Each instrument reveals a different kind of structure in the
  search landscape, even though the distant objective remains unresolved.}
  \label{fig:observable-progress}
\end{figure}

Final outcomes provide too little feedback to direct an expensive search. Most fuzzing executions do not expose a bug, yet branch and edge coverage reveal which program behavior each input reached. This intermediate signal lets the campaign observe progress before it finds a failure. Auto-research needs an analogous signal because most experiments do not establish a scientific finding, and stronger validation may arrive only after replication, expert review, or a physical assay. The signal should measure how an experiment changes the state of the search, not estimate whether the current candidate is already a discovery. In \autoref{fig:observable-progress}, we illustrate this distinction: instruments reveal structure in the search landscape while the distant finding remains uncertified.

Coverage gives fuzzers one answer to the resulting design question: what intermediate metric can expose useful progress before the final outcome is known? Research needs its own answer. A program and its instrumentation define a fixed but usually unenumerated set of observable features, and a campaign records which features it has reached. A research problem is less stable because its hypotheses and representations can change. The analogy therefore supplies three requirements for a useful progress signal, rather than a ready-made metric:

\begin{enumerate}[label=\textbf{C\arabic*.},leftmargin=*,itemsep=1pt,topsep=3pt]
  \item \textbf{Cheap and informative on each candidate}. A fuzzer evaluates coverage and sanitizer checks during the same execution, but the oracle usually returns only ``no detected failure.'' That result provides little direction for the next input. Guidance must instead report partial progress on every candidate. In research, replication, an assay, or expert review may be purchased separately, so this reading should also cost substantially less than the final validation that it helps allocate.
  \item \textbf{Correlated with genuine progress}. Following the signal should lead toward real discoveries rather than toward larger values of the signal alone.
  \item \textbf{Robust under optimization}. A search can game any signal that it maximizes \citep{gao2023scaling,skalse2022defining}.
\end{enumerate}

These requirements respond to known failures. Curiosity driven by prediction error can fixate on a ``noisy television'' and violate C3 \citep{burda2019exploration,pathak2017curiosity}. Coverage can also overcount progress: many coverage-unique crashes may collapse to a small number of distinct bugs \citep{klees2018evaluating}.

For research, the analogue of coverage should represent \emph{epistemic progress}: whether an experiment reduces uncertainty about which hypothesis holds. Such an experiment may rule out a competing explanation, locate a boundary, sharpen a held-out prediction, or change which experiment the agent runs next. C2 asks a proxy to track this progress. A narrow experiment that separates two hypotheses should score highly, while a surprising result that resolves nothing should score poorly. The signal should also remain cheaper and less model intensive than full Bayesian information gain, which requires an explicit likelihood, prior, design domain, and utility. Candidate proxies include \emph{surprise against a model}, \emph{novelty against prior work}, \emph{verifier confidence and cross-replicate disagreement}, and \emph{distance to a stated target}. Each balances C1 through C3 differently.

Researchers must test whether any proxy tracks real discovery. Fuzzing provides a suitable evaluation pattern. Across many programs, coverage correlated with bug finding, but the fuzzer with the most coverage did not necessarily find the most bugs \citep{bohme2022reliability}. A research evaluation can instrument a fixed loop with competing proxies, run repeated trials, and measure how well each proxy predicts protected outcomes and allocates a fixed validation budget. It can then compare the number of validated discoveries per unit cost. A useful proxy must survive both tests.

This capability claim makes a falsifiable prediction. \textbf{Signal prediction:} with the proposer, task, and candidate pool fixed, a useful intermediate metric should predict protected outcomes and select more validator-confirmed discoveries under a fixed validation budget than an uninformative or terminal-only signal. Whether a selected candidate counts as a discovery is a distinct final-validation question addressed in \autoref{sec:decoupling}. If no candidate signal predicts protected progress, this claim is wrong.

\section{Lesson Two: Use Feedback to Search}
\label{sec:search}

\begin{figure*}[!t]
  \centering
  \includegraphics[width=.8\linewidth]{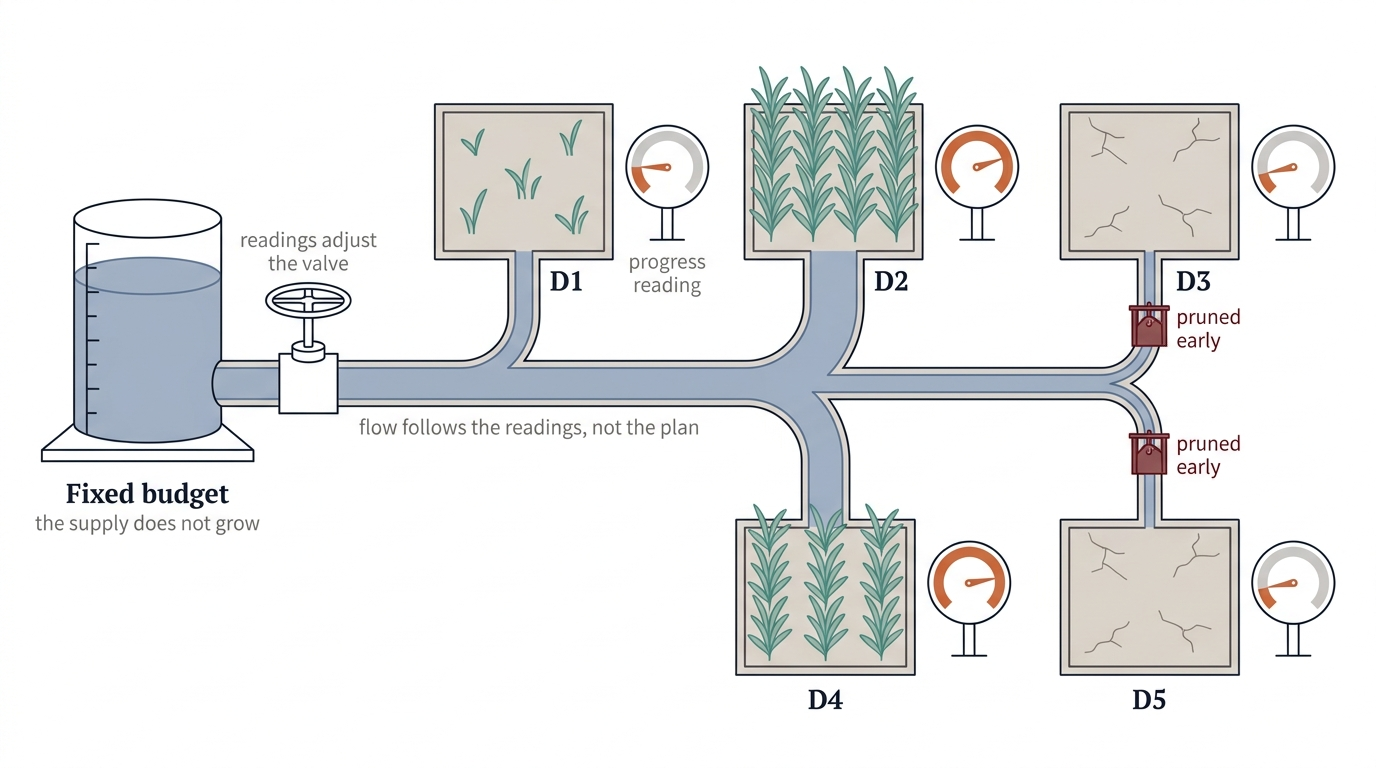}
  \caption{A scheduler concentrates a fixed budget on the directions whose
  progress readings improve, and prunes the rest early. Because the supply does
  not grow, pruning D3 and D5 frees flow for D2. Directed greybox fuzzers
  schedule energy the same way, by distance to a target \citep{aflfast,aflgo}.}
  \label{fig:feedback-search}
\end{figure*}

An intermediate signal creates value only when the search policy acts on it. A common generate-and-rank baseline prompts a proposer for $k$ candidates, scores each terminal output, and keeps the best. A tree or tournament may organize the candidates. Yet the procedure remains close to repeated sampling when experimental feedback determines only which complete sample survives, rather than which intervention the agent attempts next. In \autoref{fig:feedback-search}, we visualize how observations route a fixed resource budget among research directions.

Fuzzers use feedback more directly. AFL and AFL++ retain random mutation but allocate future mutations according to execution history \citep{afl,fioraldi2020aflpp}. Angora adds targeted constraint solving: taint analysis identifies relevant input bytes, a branch predicate becomes a black-box function $f(x)$, and a numerical approximation to $\nabla f$ guides the input until the branch flips \citep{chen2018angora}. In the relevant ablation, gradient descent solved more branch constraints than random mutation on every tested program when the input corpus was fixed. On \texttt{xmlwf}, for example, gradient descent solved 97.0\% of branch constraints, compared with 77.4\% for random mutation. Directed greybox fuzzing applies the same idea to scheduling by concentrating energy according to distance from a target \citep{aflgo}.

The lesson is to let observations determine the next transformation or allocation, not to use gradients for every scientific hypothesis. A research agent could use sensitivity analysis to identify the premise that drives an outcome, schedule compute toward under-explored directions, or generate an experiment that distinguishes the surviving explanations. AlphaEvolve and ALE-Agent already exploit executable scores in iterative search \citep{novikov2025alphaevolve,imajuku2025alebench}. Language models also act as coverage-guided mutators inside software fuzzing loops \citep{zhang2026llamafuzz,lyu2024promptfuzz}. We need to determine which operators improve protected, cost-normalized scientific outcomes rather than only the steering proxy.

Research agents will not inherit Angora's gains automatically. \textbf{Search prediction:} with the proposer, task, protected validator, and total budget fixed, a policy that uses intermediate experimental feedback to choose its next intervention should yield more validator-confirmed discoveries per unit cost than repeated sampling. This comparison defines \emph{principled search} operationally: the next action depends on the observed campaign history, and success is measured under a common cost and validation rule.

\section{Lesson Three: Separate Exploration Signals from Final Validation}
\label{sec:decoupling}

The progress signal tells the agent where to search next. Final validation decides whether the selected candidate counts as a discovery. Because the search adaptively optimizes the progress signal, that final verdict must rely on evidence protected from repeated use during search. The two roles differ in what they answer, not in when they run or what they cost.

Standard fuzzing terminology already separates access levels: black-box fuzzers observe external behavior, white-box fuzzers use detailed program analysis, and greybox fuzzers collect lightweight execution feedback \citep{manes2021fuzzing}. We extend this access-based spectrum rather than redefine it. Coverage counters and sanitizer checks may run during the same instrumented execution, but their outputs answer different questions: guidance asks ``where should the campaign search next?'', while the research analogue of an oracle asks ``does this candidate count as a discovery?''

The AI co-scientist makes this distinction concrete \citep{gottweis2025coscientist}. Its store of literature and hypotheses supplies seeds. Debate and evolution generate variants. An LLM tournament assigns Elo ratings that determine which hypotheses receive further development. Selected biomedical hypotheses eventually undergo experimental validation. In the fuzzing analogy, Elo provides the dense guidance proxy, while the physical experiment provides the final verdict. Elo can organize the search only when it predicts later correctness and experimental value. An Elo gain does not establish that the hypothesis is true.

This example shows why guidance is not a verdict. Ranking hypotheses by Elo guides search. Treating Elo as proof of scientific validity would certify a result. Search can use Elo despite its exposure to adaptive optimization, but certification requires evidence protected from that optimization, such as a hidden test, replication, physical assay, or independent review. The two signals need not run at different frequencies or have inherently different costs. The requirement is that gains in the steering proxy remain claims awaiting final validation.

\paragraph{Where final validation enters.}
Other systems place the boundary elsewhere. FunSearch rejects invalid programs and scores valid ones with a deterministic evaluator \citep{romeraparedes2024funsearch}; AlphaEvolve combines machine-executable validity checks with machine-executable objectives \citep{novikov2025alphaevolve}. Both operate in domains where systems can compute correctness and progress comparatively easily. The AI Scientist-v2 uses learned evaluators to select experimental nodes. Humans choose its initial ideas and completed manuscripts for submission, and workshop reviewers provide the eventual acceptance judgment \citep{yamada2025aiscientistv2}. These systems fall at different points on a spectrum. The relevant distinction is how much adaptive search occurs before evidence independent of the search proxy enters the loop.

The MLE-bench analysis provides a quantitative warning. Agents search with cross-validation scores and return the validation-best solution. Counterfactually choosing the test-best node would raise the medal rate by 9 to 13 percentage points \citep{toledo2025mlebench}. Researchers cannot turn the hidden test into another repeatedly queried steering score, but the gap shows that adaptive progress on the proxy does not guarantee progress under protected evaluation.

The three roles also make failures diagnosable. A failed campaign can reflect uninformative guidance, a search policy that missed the relevant region, or a final validator that miscertified the result. A conflated score obscures these explanations and leaves retraining the judge as the only apparent remedy. Fuzzing's reliability studies illustrate the alternative \citep{bohme2022reliability,klees2018evaluating}: each component has its own instrumentation and track record, so practitioners can identify which one failed.

\paragraph{What final validation buys.}
Final validation limits adaptive self-deception. A search that optimizes and reports one learned score can exploit misspecification in that score \citep{gao2023scaling,skalse2022defining,pan2022effects}. A final validator cannot repair a poor proxy, and the campaign may still waste its budget. It can, however, prevent proxy gains from automatically becoming claims of discovery. We therefore predict an epistemic benefit: fewer false discoveries among reported results. Operationally, with the candidates and search trace fixed, evidence protected from adaptive reuse should reduce the false-discovery rate relative to reporting the optimized guidance score. This is a reliability claim about the final verdict, not another search capability.

\section{A Worked Example: Instrumenting a Physics-Discovery Agent}
\label{sec:example}

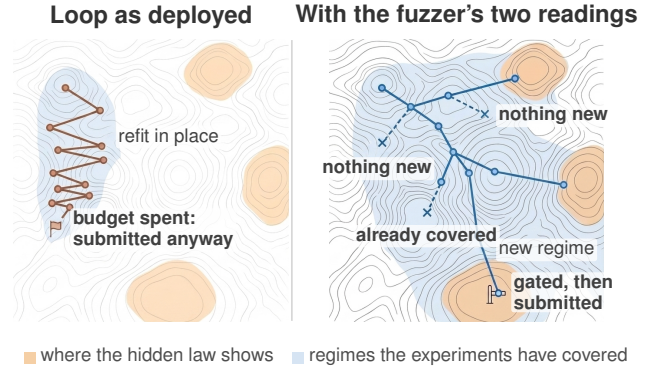
\begin{figure}[t]
  \centering
  \definecolor{tsorange}{HTML}{F5CFA5}
  \definecolor{tshalo}{HTML}{D6E4F3}
  \tikzset{
    tsmap/.style={inner sep=0, anchor=south west},
    tstitle/.style={font=\sffamily\bfseries\small, text=black!80,
      anchor=south, inner sep=0},
    tslabel/.style={font=\sffamily\bfseries\scriptsize, text=black!75,
      fill=white, fill opacity=0.75, text opacity=1, inner sep=1pt,
      rounded corners=1pt, align=left},
    tsnote/.style={tslabel, font=\sffamily\scriptsize},
  }
  \newcommand{\tspanel}[2]{%
    \begin{tikzpicture}
      \node[tsmap] (m) {\includegraphics[width=0.44\linewidth]{#1}};
      \begin{scope}[x={(m.south east)}, y={(m.north west)}]
        #2
      \end{scope}
    \end{tikzpicture}}
  \tspanel{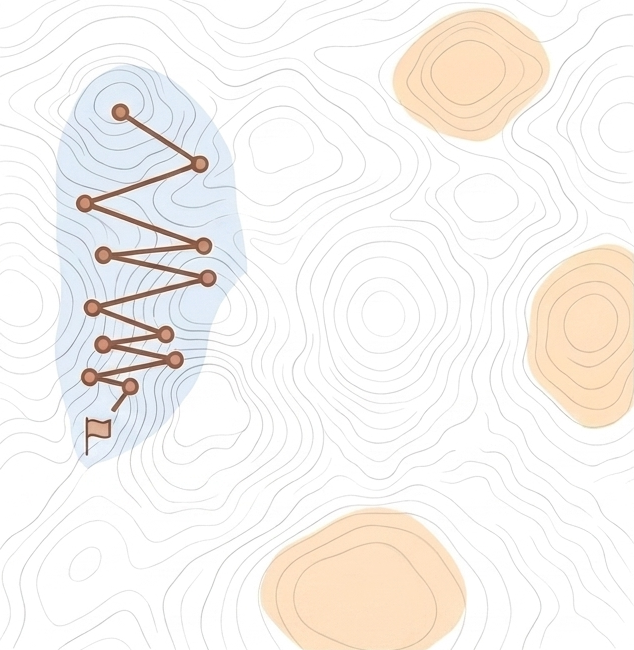}{%
    \node[tstitle] at (0.5,1.04) {Loop as deployed};
    \node[tsnote, anchor=west] at (0.37,0.655) {refit in place};
    \node[tslabel, anchor=west] at (0.205,0.325)
      {budget spent:\\submitted anyway};
  }\hfill
  \tikz\draw[black!25, line width=0.4pt] (0,0) -- (0,0.45\linewidth);\hfill
  \tspanel{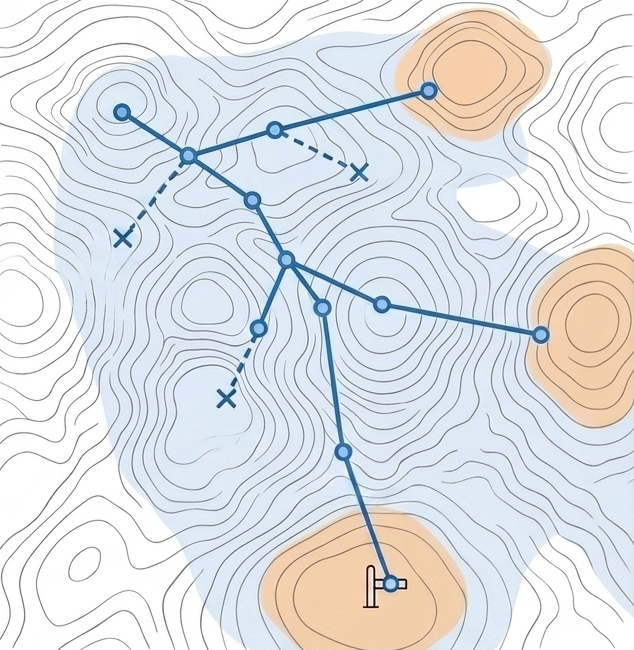}{%
    \node[tstitle] at (0.5,1.04) {With the fuzzer's two readings};
    \node[tslabel, anchor=north] at (0.175,0.585) {nothing new};
    \node[tslabel, anchor=west] at (0.605,0.73) {nothing new};
    \node[tslabel, anchor=north] at (0.355,0.345) {already covered};
    \node[tsnote, anchor=west] at (0.605,0.255) {new regime};
    \node[tslabel, anchor=west] at (0.66,0.105) {gated, then\\submitted};
  }\\[3pt]
  {\sffamily\scriptsize\color{black!70}%
    \tikz[baseline=-0.5ex]\fill[tsorange] (0,-0.8ex) rectangle (1.2ex,0.4ex);\
    where the hidden law shows\quad
    \tikz[baseline=-0.5ex]\fill[tshalo] (0,-0.8ex) rectangle (1.2ex,0.4ex);\
    regimes the experiments have covered}
  \caption{One budget, two searches. Same hidden terrain, twelve experiments
  each; orange marks where the hidden law shows, blue the regimes covered.
  Left: the loop as deployed refits in place and submits when the budget runs
  out. Right: the fuzzer's two readings prune what reveals nothing, move into
  new regimes, and gate the final law. Schematic, not run data.}
  \label{fig:two-searches}
\end{figure}

A fuzzer asks two questions after every execution: Did anything
new happen? Where have I not been? Both are cheap to answer within a research
loop, and both can change what an agent discovers. We added both readings to
an agent on DiscoverPhysics, a benchmark of simulated physics worlds. In each
round, the agent designs an experiment, observes noisy trajectories, and fits
candidate laws. It ultimately submits one law as code with a mechanistic
explanation. The benchmark grades the submission on held-out test cases and a
judged explanation. The agent sees neither, so the verdict remains protected
in the sense of \autoref{sec:decoupling}. We report one worked example as an
existence proof, not an evaluation.

We replayed 31 archived runs produced by three frontier-model
agents and found two baseline failure patterns. In the first, the submitted
law failed to explain the agent's own data: one final candidate sat at
$3.7\sigma$ against the observations the agent had collected, yet the agent
submitted it. In the second, the submitted law explained everything the agent
saw but remained wrong: one agent probed only tame regimes where a wrong
mechanism matches the right one, reproduced every held-out trajectory within
noise, and named the wrong force law. The first agent never asked whether
anything in its data was new; the second never asked where it had not been.
We contrast the searches produced by the
same budget with and without the two questions (\autoref{fig:two-searches}).

We instrument the agent as a fuzzer instruments a program. The
same design applies beyond physics. The loop has three parts. First, it
computes readings that answer the two questions above using only artifacts the
agent has already produced. Second, it injects those readings into the control
loop so that they can influence the next action. Third, it keeps certification
outside the loop, as \autoref{sec:decoupling} requires. For DiscoverPhysics,
our task harness computes two readings.\footnote{We release the harness as an
open agent skill:
\url{https://github.com/ARA-Labs/Agent-Native-Research-Artifact/tree/main/skills/research-fuzzer}.}
\emph{Surprise} is the $\sigma$-normalized residual of the current law on the
data collected so far; the \emph{envelope} records the range explored for each
experimental knob. After each experiment, the panel presents both readings,
and the agent must acknowledge them in one sentence before acting. Placement
matters: the same readings at the top of the prompt went unmentioned for an
entire run. No protected information crosses the boundary; held-out tests and
the judged explanation stay outside the loop.

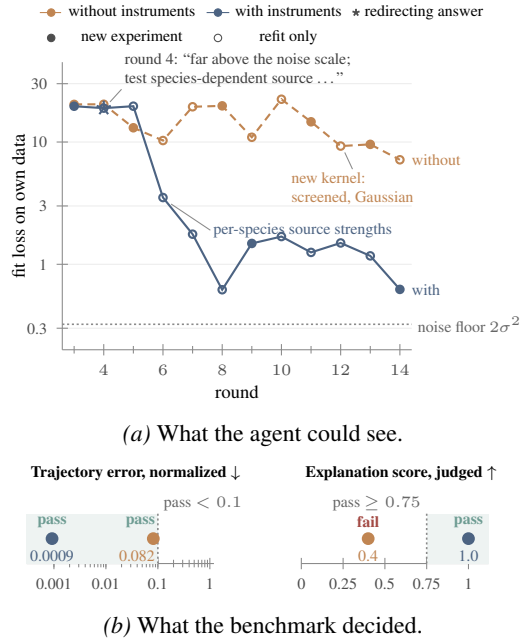
\begin{figure}[t]
  \centering
  \definecolor{runoff}{HTML}{C08552}
  \definecolor{runon}{HTML}{4C6382}
  \definecolor{passteal}{HTML}{6E9C93}
  \definecolor{oxblood}{HTML}{A34A44}
  \definecolor{axisgrey}{HTML}{8A8A8A}
  \pgfplotsset{
    tsaxis/.style={
      scale only axis,
      axis lines=left, axis line style={-, axisgrey, line width=0.4pt},
      tick style={axisgrey, line width=0.4pt},
      tick label style={font=\tiny, text=black!70},
      label style={font=\scriptsize},
      every axis plot/.append style={line width=0.8pt},
      clip=false,
    },
    tsoff/.style={runoff, mark=*, mark size=1.4pt,
      mark options={fill=runoff, draw=runoff, solid}},
    tson/.style={runon, mark=*, mark size=1.4pt,
      mark options={fill=runon, draw=runon}},
    refline/.style={axisgrey, densely dotted, line width=0.6pt},
    verdict/.style={tsaxis, width=0.30\linewidth, height=0.26in,
      ymin=0, ymax=1, ytick=\empty, axis y line=none,
      title style={font=\tiny\bfseries, anchor=south west,
        at={(0,1)}, xshift=-2pt, yshift=3pt}},
  }
  \tikzset{
    note/.style={font=\tiny, text=black!75, inner sep=1pt, align=left},
    leader/.style={axisgrey, line width=0.4pt, shorten >=1pt},
    tag/.style={font=\tiny\bfseries, inner sep=1pt},
  }
  \begin{tikzpicture}
    \begin{axis}[hide axis, xmin=0, xmax=1, ymin=0, ymax=1,
      scale only axis, width=1cm, height=1cm,
      legend columns=3, legend to name=leg:three-species,
      legend style={font=\tiny, draw=none, fill=none,
        /tikz/every even column/.append style={column sep=2pt}},
      legend image code/.code={\draw[mark repeat=2, mark phase=2, #1]
        plot coordinates {(0cm,0cm) (0.15cm,0cm) (0.30cm,0cm)};},
      legend image post style={mark size=1.4pt}]
      \addlegendimage{tsoff, densely dashed}
      \addlegendentry{without instruments}
      \addlegendimage{tson}
      \addlegendentry{with instruments}
      \addlegendimage{black!70, only marks, mark=star, mark size=2.2pt,
        line width=0.7pt}
      \addlegendentry{redirecting answer}
      \addlegendimage{black!70, only marks, mark=*}
      \addlegendentry{new experiment}
      \addlegendimage{black!70, only marks, mark=o}
      \addlegendentry{refit only}
    \end{axis}
  \end{tikzpicture}%
  \pgfplotslegendfromname{leg:three-species}\\[-3pt]
  \begin{subfigure}[t]{\linewidth}
    \centering
    \begin{tikzpicture}
      \begin{axis}[tsaxis, width=0.56\linewidth, height=1.5in,
        ymode=log, xmin=2.6, xmax=14.4, ymin=0.2, ymax=45,
        xtick={4,6,8,10,12,14}, minor xtick={3,5,...,13},
        ytick={0.3,1,3,10,30}, yticklabels={0.3,1,3,10,30},
        xlabel={round}, ylabel={fit loss on own data},
        xlabel style={yshift=3pt}, ylabel style={yshift=-6pt},
        ymajorgrids, grid style={black!8, line width=0.3pt},
        log ticks with fixed point,
        ]
        \addplot[refline, domain=2.6:14.4] {0.3231}
          node[pos=1, note, anchor=west, xshift=1pt, text=black!60]
          {noise floor $2\sigma^2$};
        \addplot[tsoff, densely dashed, mark=none]
          table[x=round, y=loss] {figures/data/three-species-off.dat};
        \addplot[tsoff, only marks]
          table[x=round, y=loss, restrict expr to domain={\thisrow{newexp}}{1:1}]
          {figures/data/three-species-off.dat};
        \addplot[tsoff, only marks, mark=o, mark options={fill=white}]
          table[x=round, y=loss, restrict expr to domain={\thisrow{newexp}}{0:0}]
          {figures/data/three-species-off.dat};
        \addplot[tson, mark=none]
          table[x=round, y=loss] {figures/data/three-species-on.dat};
        \addplot[tson, only marks]
          table[x=round, y=loss, restrict expr to domain={\thisrow{newexp}}{1:1}]
          {figures/data/three-species-on.dat};
        \addplot[tson, only marks, mark=o, mark options={fill=white}]
          table[x=round, y=loss, restrict expr to domain={\thisrow{newexp}}{0:0}]
          {figures/data/three-species-on.dat};
        \addplot[runon, only marks, mark=star, mark size=2.8pt, line width=0.8pt]
          coordinates {(4,18.9299)};
        \node[note, text=runoff, anchor=west] at (axis cs:14.3,7.157) {without};
        \node[note, text=runon, anchor=west] at (axis cs:14.3,0.623) {with};
        \node[note, anchor=south west] (ack) at (axis cs:4.6,27)
          {round 4: ``far above the noise scale;\\
           test species-dependent source \ldots''};
        \draw[leader] (ack.south west) -- (axis cs:4.15,20.5);
        \node[note, anchor=west, text=runon] (src) at (axis cs:7.6,1.9)
          {per-species source strengths};
        \draw[leader] (src.west) -- (axis cs:6.2,3.2);
        \node[note, anchor=north, text=runoff] (ker) at (axis cs:12.4,6.0)
          {new kernel:\\screened, Gaussian};
        \draw[leader] (ker.north) -- (axis cs:12.1,8.6);
      \end{axis}
    \end{tikzpicture}
    \caption{What the agent could see.}
    \label{fig:three-species-a}
  \end{subfigure}\\[2pt]
  \begin{subfigure}[t]{\linewidth}
    \centering
    \begin{tikzpicture}
      \begin{axis}[verdict, name=traj,
        xmode=log, xmin=3e-4, xmax=1.2,
        xtick={0.001,0.01,0.1,1}, xticklabels={0.001,0.01,0.1,1},
        title={Trajectory error, normalized $\downarrow$},
        ]
        \fill[passteal!12] (axis cs:3e-4,0) rectangle (axis cs:0.1,1);
        \addplot[refline] coordinates {(0.1,0) (0.1,1)}
          node[pos=1, note, anchor=south west, xshift=1pt, text=black!60]
          {pass $<0.1$};
        \addplot[tson, only marks, mark size=2pt] coordinates {(0.00092,0.5)}
          node[note, anchor=north, yshift=-3pt, text=runon] {0.0009}
          node[tag, anchor=south, yshift=3pt, text=passteal] {pass};
        \addplot[tsoff, only marks, mark size=2pt] coordinates {(0.0817,0.5)}
          node[note, anchor=north east, yshift=-3pt, xshift=2pt, text=runoff] {0.082}
          node[tag, anchor=south east, yshift=3pt, xshift=2pt, text=passteal] {pass};
      \end{axis}
      \begin{axis}[verdict, at={(traj.east)}, anchor=west,
        xshift=0.14\linewidth,
        xmin=0, xmax=1.12,
        xtick={0,0.25,0.5,0.75,1}, xticklabels={0,0.25,0.5,0.75,1},
        title={Explanation score, judged $\uparrow$},
        ]
        \fill[passteal!12] (axis cs:0.75,0) rectangle (axis cs:1.12,1);
        \addplot[refline] coordinates {(0.75,0) (0.75,1)}
          node[pos=1, note, anchor=south east, xshift=-1pt, text=black!60]
          {pass $\geq 0.75$};
        \addplot[tsoff, only marks, mark size=2pt] coordinates {(0.4,0.5)}
          node[note, anchor=north, yshift=-3pt, text=runoff] {0.4}
          node[tag, anchor=south, yshift=3pt, text=oxblood] {fail};
        \addplot[tson, only marks, mark size=2pt] coordinates {(1.0,0.5)}
          node[note, anchor=north, yshift=-3pt, text=runon] {1.0}
          node[tag, anchor=south, yshift=3pt, text=passteal] {pass};
      \end{axis}
    \end{tikzpicture}
    \caption{What the benchmark decided.}
    \label{fig:three-species-b}
  \end{subfigure}
  \caption{One world, two runs of the same agent. (a) Fit loss on
  the agent's own data against the noise floor a correct law reaches; the
  starred round-four answer redirected the search (the panel was acknowledged
  every round). (b) The protected verdict neither run could see.}
  \label{fig:three-species}
\end{figure}

In \autoref{fig:three-species}, we compare two runs and trace how the panel changed the verdict. In a
world with three hidden particle species, the agent fit twelve successive laws
in both runs. Without the panel, the loss never came within twenty times the
noise floor the agent had measured. The submitted law also placed the species
in a field of the wrong form. Its trajectory error fell below the benchmark's
threshold, but the judged explanation scored 0.4 against a threshold of 0.75.
With the panel, the agent reached the same plateau. In round four, it
responded: ``the valid-case fit is far above the noise scale; I will therefore
test whether the missing structure is species-dependent source and response
charge rather than collect more redundant trajectories.'' Subsequent fits
introduced per-species source strengths and reduced the loss from 19.6 to
about twice the noise floor. The submitted law named the three species in the
ratio $1:3:-2$ and scored 1.0 on the explanation. Its trajectory error was
smaller by a factor of ninety.

\section{Related Work}
\label{sec:related}

Many systems now automate parts of research, and recent surveys identify evaluation as a weak point \citep{eger2025transforming,reddy2025towards}. Researchers debate whether static benchmarks, deployment evidence, or verification inside the loop should measure progress \citep{chen2026summit}. Jerry Tworek argues that ``the era of evals is done'': as agents improve at generating research outputs, the bottleneck moves from generation to verification, and sustained improvement in deployed workflows replaces static benchmarks as the meaningful test \citep{agihouse2026tworek}. This shift names the same gap we formalize: a learned judge cannot substitute for independent validation. We add three questions to that debate: what intermediate signal makes progress observable, how does the search policy act on it, and what independent evidence validates the output? Existing systems occupy intermediate positions along these dimensions.

Ninety years of work on mechanized verification and self-improving
systems rehearse the same failure points. Hilbert asked for a mechanical
procedure that decides whether a first-order formula is
valid; \citet{godel1931unentscheidbare} showed that strong formal systems cannot prove all truths
about themselves, \citet{church1936unsolvable} and \citet{turing1936computable} answered the Entscheidungsproblem in the
negative, and \citet{rice1953classes} extended the verdict to every nontrivial property of
program behavior.
Because verification resists full mechanization, software practice retreated
to empirical search that pairs cheap guidance with partial oracles.
Self-improving programs kept rediscovering this boundary. \citet{good1966ultraintelligent} speculated
that an ultraintelligent machine could redesign itself; \citet{friedberg1958learning} mutated machine code at random
and found that unguided sampling learned no faster than exhaustive random
search. The G\"odel machine of \citet{schmidhuber2009godel} accepted
only provably beneficial self-modifications, and the cost of proof kept it
from taking a single step: a protected verdict without a cheap signal. Neural architecture search steered and
certified with one static benchmark \citep{zoph2017nas}, and the search
overfit the benchmark it optimized \citep{yang2020naseval}. Learned research judges now
repeat that conflation at the scale of papers. Each wave stalled where the
fuzzing loop predicts: guidance was missing, the oracle was unaffordable, or
one score played both roles. Science likewise has no procedure that decides
whether a hypothesis is true, so auto-research should inherit the working
answer rather than await a perfect verifier.

The AI Scientist family and AI co-scientist use the learned evaluation mechanisms analyzed in \autoref{sec:decoupling}, but external review or experiments ultimately judge selected outputs \citep{lu2024aiscientist,yamada2025aiscientistv2,gottweis2025coscientist}. Agent Laboratory likewise ranks code with a generated reward and grades manuscripts with an emulated reviewer \citep{schmidgall2025agentlab}. These signals can organize search, but their scores do not provide independent scientific validation. MLR-Bench illustrates the risk under its coding-agent conditions: roughly 80\% of cases contain fabricated or invalidated experimental results \citep{chen2025mlrbench}.

Systems with executable objectives face an easier validation problem. FunSearch and AlphaEvolve reject invalid programs and score valid ones with machine-executable evaluators \citep{romeraparedes2024funsearch,novikov2025alphaevolve}. ALE-Agent iteratively optimizes an executable contest score and placed 21st among more than \num{1000} participants in a live AtCoder contest \citep{imajuku2025alebench}. In the wet lab, Coscientist can execute selected chemistry procedures \citep{boiko2023coscientist}, and the Virtual Lab sent its final nanobody designs to physical assays \citep{swanson2025virtuallab}. These examples are consistent with the value of grounded evaluation, but they span incomparable tasks and cannot isolate the effect of observability, search, or validation design.

Two lines of work support this framing. AIGS adds a dedicated falsification agent that separates verification from generation \citep{liu2024aigs}; Dolphin filters ideas with cheap LLM signals but advances only those that improve an executed task metric \citep{yuan2025dolphin}. Separately, LLM judges favor their own generations, change rankings under reordering, and respond to injected content in scholarly review \citep{zheng2023judging,panickssery2024llm,wang2024large,ye2024arewethereyet}. Results on reward overoptimization and adaptive validity explain how repeated optimization can amplify these weaknesses \citep{gao2023scaling,skalse2022defining,pan2022effects,dwork2015reusable}. Evaluation harnesses such as RE-Bench and MLAgentBench already separate a running signal from a final executable score \citep{wijk2024rebench,huang2024mlagentbench}. We frame these choices as observability, search, and validation design, then use fuzzing to derive controlled comparisons among them. Content fuzzing already demonstrates that the architecture transfers beyond software testing, though in a narrow domain \citep{he2026contentfuzz}; open-ended science still needs progress signals that satisfy the criteria in \autoref{sec:signal}.

Agent-native research artifacts supply the surface this framing requires. \citet{liu2026humanwrittenpaperagentnativeresearch} propose that research outputs ship as machine-executable artifacts rather than prose papers, so agents can parse, run, and extend them directly. Such artifacts open research to analysis the way instrumented binaries open programs: dynamic analysis, such as the fuzzing loop we describe, executes them and observes feedback, while static analysis, such as verification and interpretation, inspects their claims without execution. Our proposals for dense progress signals and protected validation presuppose this surface; artifact-native outputs make it available.

\section{Alternative Views}
\label{sec:alternatives}

Six objections define the limits of our position.

\paragraph{``LLMs cannot make the abductive jump that discovery requires.''}
\citet{zahavy2026jump} argues that current LLMs cannot move from sensory experience to genuinely new axioms through induction or deduction alone. Our claim begins after a research problem and candidate representation exist. It asks how an agent should observe progress, search, and validate within them. Interactive world models may expand what an agent can propose; fuzzing-style feedback can then make the downstream search more efficient and its discoveries more trustworthy. Our analogy does not explain how an agent changes what counts as the problem, a candidate, or a discovery.

\paragraph{``This is Bayesian optimal experimental design renamed.''}
BOED occupies the model-rich end of our proposed research spectrum \citep{lindley1956measure,rainforth2024modern}. It requires an explicit probabilistic model, design domain, and utility. The design domain may be continuous, and modern estimators can reduce the cost of expected information gain. Our proposal addresses settings in which researchers cannot yet specify these ingredients reliably. In such settings, a cheaper empirical proxy can guide exploration while replication, held-out prediction, or expert review provides stronger validation. BOED offers a principled solution when its model is credible; fuzzing is a general feedback-guided, mutation-based paradigm that generalizes beyond software to other objects of analysis~\citep{he2026contentfuzz}.

\paragraph{``This is novelty search or quality-diversity with extra steps.''}
The approaches are related \citep{lehman2011abandoning,mouret2015illuminating}, but they differ from fuzzing in what the guidance signal measures. A quality-diversity behavior descriptor is hand-designed to describe progress toward the goal; coverage is read automatically off the candidate's own execution and describes what it did, not how close it came. The signal is a property of the process, not an estimate of the outcome, so it needs no model of what counts as a discovery. That property is what open research requires, because the outcome is exactly what cannot yet be modeled. Novelty search and Bayesian acquisition, by contrast, both score a candidate through a model or descriptor of the goal. Quality-diversity also folds steering and acceptance into one descriptor, whereas we keep a protected verdict separate.

\paragraph{``The bitter lesson says scale the generator, not engineer the feedback.''}
The bitter lesson argues for search methods that can exploit computation, rather than for any particular generator \citep{sutton2019bitter}. Fuzzing follows that principle when it uses general execution feedback instead of hand-coded domain rules. Repeated sampling also uses computation and can work well, so the available evidence does not justify rejecting it universally. Across the benchmarks studied by \citet{brown2024monkeys}, however, the fraction of problems solved grew approximately log-linearly over the measured range and produced diminishing marginal returns. We claim that adaptive feedback can improve this exchange rate without replacing scalable computation with hand-tuned heuristics.

\paragraph{``Open-ended research has no stable search space, so coverage is undefined.''}
Coverage does not require an enumeration of every reachable behavior, but a fixed target and its instrumentation still define what the fuzzer can observe. Research is harder because its hypotheses and representations may change during a campaign. Our proposal therefore assumes a declared problem and an explicit progress representation that can grow as observations accumulate. It does not cover an agent that redefines what counts as a candidate or discovery. Within this boundary, guidance can compare each new experiment with the accumulated record rather than with an enumerated whole.

\paragraph{``Research has no oracle, so the analogy breaks.''}
Fuzzing has its own oracle problem \citep{barr2015oracle}, and real campaigns use imperfect, deferred, differential, or human-assisted checks. A typical campaign does not separate oracle and guidance by invocation schedule or assume that sanitizer checks inherently cost more. The same instrumented execution updates coverage and checks relevant operations for violations. The signals differ in role: the oracle judges the current input, while guidance identifies behavior absent from the campaign history. Guidance cannot certify a bug, and the oracle cannot efficiently explore the input space. Scientific validation may also be expensive or partial, so this semantic separation helps allocate validation to selected candidates.

\section{Conclusion}

Auto-research should operate as instrumented, feedback-directed search: expose a cheap measure of epistemic progress; let the search policy use that measure to choose the next intervention; and let protected validation decide what counts as a discovery, since an optimized progress signal can no more certify a finding than coverage can certify a bug. The position stands or falls on the three controlled comparisons stated above: candidate signals must predict protected outcomes, feedback-directed search must beat repeated sampling per validated discovery, and protected validation must reduce false discoveries relative to the optimized proxy. This division also locates the durable human role: researchers declare the problem at the front and ground the verdict at the back, while the agent runs the loop between them. Beneath these three lessons, observable progress, principled search, and protected validation, sits one belief: a scientific experiment, like a program, can offer intermediate guidance that exposes progress before its outcome is known. Whether it can is the open question we hand the field.

\printbibliography

\end{document}